\pdfoutput=1
\documentclass[11pt]{article}

\usepackage[final]{acl}
\usepackage[most]{tcolorbox}
\usepackage{times}
\usepackage{float}
\usepackage{multirow}
\usepackage{enumitem}
\usepackage{pifont}
\usepackage{subfig}
\usepackage{subcaption}
\usepackage{subfig}
\usepackage{latexsym}

\usepackage[T1]{fontenc}
\usepackage[utf8]{inputenc}
\usepackage{booktabs}
\usepackage{microtype}

\usepackage{inconsolata}

\usepackage{graphicx}

\title{VaxGuard: A Multi-Generator, Multi-Type, and Multi-Role Dataset for Detecting LLM-Generated Vaccine Misinformation}

\author{
 \textbf{Syed Talal Ahmad\textsuperscript{1}},
 \textbf{Haohui Lu\textsuperscript{2}},
 \textbf{Sidong Liu\textsuperscript{1}},
 \textbf{Annie Lau\textsuperscript{1}},\\
 \textbf{Amin Beheshti\textsuperscript{1}},
 \textbf{Mark Dras\textsuperscript{1}},
 \textbf{Usman Naseem\textsuperscript{1}},
 \\
 \textsuperscript{1}Macquarie University, Australia 
 \textsuperscript{2}University of Sydney, Australia
 \\
 \small{
   \textbf{Correspondence:} \href{mailto:usman.naseem@mq.edu.au}{usman.naseem@mq.edu.au}
 }
}

\begin{document}
\maketitle
\begin{abstract}
Recent advancements in Large Language Models (LLMs) have significantly improved text generation capabilities. However, they also present challenges, particularly in generating vaccine-related misinformation, which poses risks to public health. Despite  research on human-authored misinformation, a notable gap remains in understanding how LLMs contribute to vaccine misinformation and how best to detect it. Existing benchmarks often overlook vaccine-specific misinformation and the diverse roles of misinformation spreaders. This paper introduces \textbf{VaxGuard}, a novel dataset designed to address these challenges. VaxGuard includes vaccine-related misinformation generated by multiple LLMs and provides a comprehensive framework for detecting misinformation across various roles. Our findings show that GPT-3.5 and GPT-4o consistently outperform other LLMs in detecting misinformation, especially when dealing with subtle or emotionally charged narratives. On the other hand, PHI3 and Mistral show lower performance, struggling with precision and recall in fear-driven contexts. Additionally, detection performance tends to decline as input text length increases, indicating the need for improved methods to handle larger content. These results highlight the importance of role-specific detection strategies and suggest that VaxGuard can serve as a key resource for improving the detection of LLM-generated vaccine misinformation.
\end{abstract}

\section{Introduction}
Large Language Models (LLMs) have revolutionized the field of text generation, producing highly sophisticated and coherent content across various domains. These advancements have led to widespread applications, from creative writing to answering complex queries, and have significantly improved automation in content production tasks \cite{brown2020language}. However, with these advancements comes a critical concern: LLMs can generate misleading or false information, particularly in sensitive areas such as public health. One of the most pressing concerns is their ability to produce vaccine-related misinformation, which can erode public trust in vaccination efforts and contribute to dangerous outcomes, such as vaccine hesitancy \cite{joseph2022covid,zhou2023synthetic}.

Vaccine-related misinformation has been shown to cause significant harm by undermining public health initiatives \cite{loomba2021measuring}. Misinformation spread through online platforms, particularly social media, has the power to reach vast audiences, swaying public opinion and increasing resistance to scientifically backed vaccination campaigns. Much of the existing research on vaccine misinformation has focused on human-authored content, such as posts from anti-vaccine activists or conspiracy theorists. While this research is critical, it often overlooks a growing concern: the ability of LLMs to generate similarly misleading and manipulative content at scale \cite{di2022vaccineu,li2022classifying,hayawi2022anti}.

Furthermore, research on detecting misinformation has generally concentrated on human-generated text, with limited attention given to misinformation generated by LLMs. These existing frameworks, while valuable, lack the granularity required to address the unique challenges posed by LLMs. Misinformation detection systems have been primarily trained on datasets that do not reflect the complexities of vaccine-related content generated by LLMs, particularly across various misinformation roles, such as conspiracy theorists or fearmongers \cite{shah2024navigating}. Moreover, many current evaluation frameworks are limited to older text generators and do not adequately account for the multi-faceted roles that actors in misinformation ecosystems play, nor the capabilities of more advanced LLMs, such as GPT-3.5, GPT-4o, LLaMA3, PHI3, and Mistral, which can craft nuanced and persuasive misinformation \cite{zellers2019defending}.

Given these developments, a critical research question arise: \textit{How effectively can LLMs detect vaccine-related misinformation generated by other LLMs?} Tackling this question is crucial for developing robust misinformation detection frameworks to handle LLM-generated content.

\begin{figure}[t]
\centering
  \includegraphics[width=0.45\textwidth]{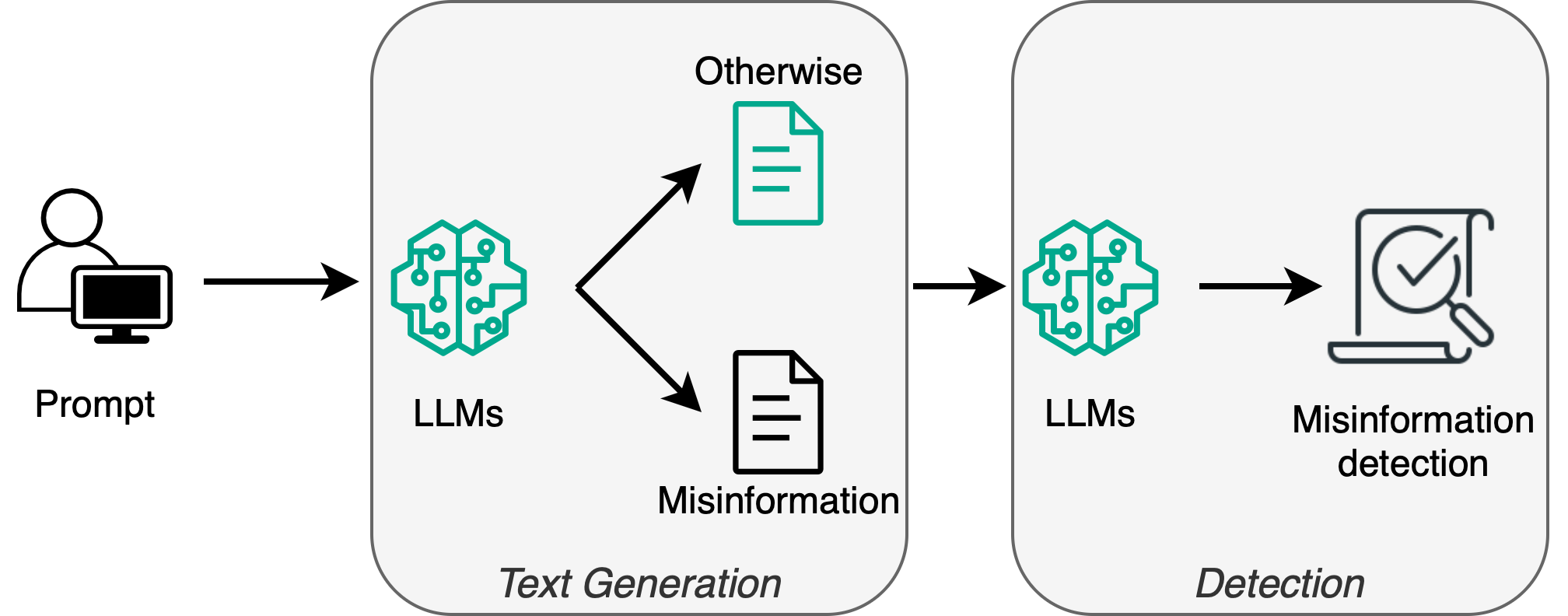}
  \vspace{-0.30cm}
  \caption{Overview of misinformation generation and detection with LLMs, showing how prompts generate "Otherwise" and "Misinformation" for misinformation detection.
  }
  \label{fig:1}
 \vspace{-0.30cm}
\end{figure}

To address these critical gaps, we introduce VaxGuard, a novel multi-generator dataset designed to detect vaccine-related misinformation generated by LLMs. VaxGuard includes misinformation content related to vaccines for COVID-19, HPV, and Influenza, created by a variety of LLMs, including GPT-3.5, GPT-4o, LLaMA3, PHI3, and Mistral. VaxGuard also spans different misinformation roles, including conspiracy theorists, fearmongers, and anti-vaccine activists. VaxGuard enables an evaluation of detection methods in terms of their ability to generalize across different LLMs and roles \cite{chen2023combating}. Figure~\ref{fig:1} shows the high-level overview of our misinformation generation and detection process.

Our analysis reveals significant challenges in the detection of LLM-generated misinformation, particularly when content is tailored to different roles. These challenges highlight the need for more sophisticated detection mechanisms that can adapt to both the evolving capabilities of LLMs and the diverse roles played by misinformation actors.

We made several key observations through comprehensive experiments. Our findings indicate that LLMs effectively generate vaccine-related misinformation, with notable similarities across different roles. Additionally, while LLMs like GPT-3.5 and GPT-4o show strong detection capabilities, others like PHI3 and Mistral face challenges with certain types of misinformation. These results emphasize the need for further refinement in detection models to handle recurring misinformation patterns and role-specific nuances more effectively. Our contributions are as follows:

% In this paper, we address this question by providing the following contributions:

\begin{itemize}[noitemsep,leftmargin=*]
% [leftmargin=*]

\item To the best of our knowledge, this is the first study specifically focusing on LLM-generated vaccine misinformation. 

\item We introduce VaxGuard, a novel dataset for role-specific detection of vaccine-related misinformation generated by multiple LLMs. 

\item We evaluate LLMs' effectiveness in detecting various types of LLM-generated misinformation, providing insights into their performance across different misinformation contexts. 
\end{itemize}

% Our work aligns with \textit{UN Sustainable Development Goal 3: Good Health and Well-Being} by detecting vaccine misinformation, supporting informed decisions, and reducing vaccine hesitancy for improved global health.

\section{Related Work}

\subsection{Vaccine Misinformation}
% \textbf{Vaccine Misinformation}: 

Vaccine misinformation has been a persistent challenge in public health, particularly as digital communication platforms have grown in influence. The rapid dissemination of false information about vaccines can undermine public health efforts, leading to reduced vaccination rates and increased outbreaks of preventable diseases. Studies have shown that vaccine misinformation can be categorized into different roles, each playing a unique part in spreading false narratives \cite{broniatowski2018weaponized, dunn2015associations, johnson2020online,naseem2024vaccine}. For example, misinformation spreaders often amplify content without verifying its accuracy, contributing to its viral nature \cite{lee2022misinformation, wilson2020social}. Religious conspiracy theorists may connect vaccines to apocalyptic or spiritual beliefs, influencing communities that hold such beliefs to resist vaccination efforts \cite{johnson2020online,naseem2021classifying}. Fear mongers exploit anxieties related to health and safety, often using emotionally charged language to sway public opinion against vaccines \cite{broniatowski2018weaponized}. Finally, anti-vacciners, who oppose vaccines on ideological grounds, contribute to the spread of misinformation by promoting discredited theories and pseudoscience~\cite{dunn2015associations}. Understanding these roles is crucial for developing strategies to combat vaccine misinformation effectively. Our work addresses this gap and presents a new dataset (VaxGuard) that spans across different roles including conspiracy theorists, fear-mongers, and anti-vaccine activists.

\subsection{Machine Generated Text}
Recent years have seen the emergence of several datasets aimed at studying machine-generated text, particularly in the context of LLMs. For instance, Med-MMHL \cite{sun2023med} provides a collection of 17,608 LLM-generated sentences focused on the medical domain, offering insights into machine-generated text across multiple diseases. LLMFake \cite{chen2023can} targets machine-generated misinformation, highlighting challenges in this space. Expanding the scope, M4GT-Bench \cite{wang2024m4gt} introduces a multilingual and multi-domain dataset with 15,000 samples generated by various LLMs, facilitating research across diverse fields. Similarly, SeqXGPT-Bench \cite{wang2023seqxgpt} includes 30,000 documents generated by models like GPT-2, GPT-3.5, and LLaMA, along with human-generated content, offering a comprehensive resource for studying machine-generated text across multiple generators. MultiSocial \cite{macko2024multisocial}, with 414,000 posts spanning 22 languages and five social media platforms, offers an extensive dataset for exploring machine-generated text from multiple platforms and languages. These datasets collectively highlight the increasing focus on machine-generated text, emphasizing the importance of benchmarking across different domains, languages, and platforms.

The rise of advanced LLMs, such as GPT-4o and LLaMA3, has raised concerns about their ability to generate convincing misinformation. While these models can produce useful content, they are also capable of creating false information that mimics human writing, making it difficult to detect \citep{chen2023can}. Furthermore, their integration into social media botnets complicates efforts to control misinformation, highlighting the need for better strategies to mitigate the spread of false information \citep{yang2023anatomy}.

\begin{table}[!t]
  \centering
  % \scriptsize  % Reduced font size to fit in one column
  \setlength{\tabcolsep}{2pt}  % Reduce space between columns
  % \caption{Summary of LLM-generated Datasets}
  % \vspace{-0.25cm}
  \scalebox{0.71}{
  \begin{tabular}{c c c c}
    \midrule \midrule
    \textbf{Dataset} & \textbf{Multi Generator} & \multicolumn{2}{c}{\textbf{Vaccine related and Roles}} \\ 
    \cmidrule{3-4}
    & & \textbf{Vaccine} & \textbf{Roles} \\ \midrule \midrule
    MGTBench \\ \citep{he2023mgtbench} & \ding{51} & \ding{55} & \ding{55} \\ \midrule
    GPABenchmark \\ \citep{liu2023check} & \ding{55} & \ding{55} & \ding{55} \\ \midrule
    MULTITuDE \\ \citep{macko2023multitude} & \ding{51} & \ding{55} & \ding{55} \\ \midrule
    M4 \\ \citep{wang2024m4} & \ding{51} & \ding{55} & \ding{55} \\ \midrule
    LLMFake \\ \citep{chen2023can} & \ding{55} & \ding{55} & \ding{55} \\ \midrule
    SeqXGPT-Bench \\ \citep{wang2023seqxgpt} & \ding{51} & \ding{55} & \ding{55} \\ \midrule
    MultiSocial \\ \citep{macko2024multisocial} & \ding{51} & \ding{55} & \ding{55} \\ \midrule
    \textbf{VaxGuard (Ours)} & \textbf{\ding{51}} & \textbf{\ding{51}} & \textbf{\ding{51}} \\ \midrule \midrule
  \end{tabular}
}
\vspace{-0.40cm}
\caption{Summary of LLM-generated Datasets}
  \label{tab:related_work}
\vspace{-0.40cm}
\end{table}

% Table~\ref{tab:related_work} compares previous LLM-generated datasets in the literature.

As shown in the Table~\ref{tab:related_work}, our study addresses a critical gap by focusing on machine-generated misinformation specifically related to vaccines. Our dataset, VaxGuard, stands out by generating misinformation from different roles in the context of various vaccines, contributing a more extensive and diverse dataset compared to prior work, making it a valuable resource for studying vaccine misinformation generated by LLMs.

\begin{table*}[!t]
\vspace{-0.10cm}
\centering
% \small
% \scriptsize
% \caption{Detailed Distribution of VaxGuard Dataset: Roles, Vaccine Types, and Generators}
% \vspace{-0.25cm}
% \label{datastats}
\scalebox{0.71}{
\begin{tabular}{l l l r r r r r r}
\midrule \midrule
\textbf{Class} & \textbf{Role} & \textbf{Vaccine Type} & \textbf{GPT-3.5} & \textbf{GPT-4o} & \textbf{LLaMA3} & \textbf{PHI3} & \textbf{Mistral} & \textbf{Total Samples} \\ 
\midrule \midrule
\multirow{12}{*}{\textbf{Misinformation}} 
& \multirow{3}{*}{Misinformation Spreader} & COVID & 500 & 500 & 500 & 500 & 500 & 2,500 \\
&  & HPV & 500 & 500 & 500 & 500 & 500 & 2,500 \\
&  & Influenza & 500 & 500 & 500 & 500 & 500 & 2,500 \\
\cmidrule(lr){3-9}
& \multirow{3}{*}{Religious Conspiracy Theorist} & COVID & 500 & 500 & 500 & 500 & 500 & 2,500 \\
&  & HPV & 500 & 500 & 500 & 500 & 500 & 2,500 \\
&  & Influenza & 500 & 500 & 500 & 500 & 500 & 2,500 \\
\cmidrule(lr){3-9}
& \multirow{3}{*}{Fear Monger} & COVID & 500 & 500 & 500 & 500 & 500 & 2,500 \\
&  & HPV & 500 & 500 & 500 & 500 & 500 & 2,500 \\
&  & Influenza & 500 & 500 & 500 & 500 & 500 & 2,500 \\
\cmidrule(lr){3-9}
& \multirow{3}{*}{Anti-Vacciner} & COVID & 500 & 500 & 500 & 500 & 500 & 2,500 \\
&  & HPV & 500 & 500 & 500 & 500 & 500 & 2,500 \\
&  & Influenza & 500 & 500 & 500 & 500 & 500 & 2,500 \\
\midrule
\multirow{3}{*}{\textbf{Otherwise}} 
& \multirow{3}{*}{N/A} & COVID & 2,000 & 2,000 & 2,000 & 2000 & 2000 & 10,000 \\
&  & HPV & 2,000 & 2,000 & 2,000 & 2000 & 2000 & 10,000 \\
&  & Influenza & 2,000 & 2,000 & 2,000 & 2000 & 2000 & 10,000 \\
\midrule
\textbf{Total} & All Roles & All Vaccine Types & 12,000 & 12,000 & 12,000 & 12,000 & 12,000 & 60,000 \\
\midrule \midrule
\end{tabular}
}
\vspace{-0.4cm}
\caption{Detailed Distribution of VaxGuard Dataset: Roles, Vaccine Types, and Generators}\label{datastats}
\vspace{-0.3cm}
\end{table*}

% \textbf{
\subsection{Use LLM-Detected Misinformation}

Using LLMs to detect misinformation, particularly the misinformation they generate, is a promising approach. Their advanced language processing capabilities allow them to analyze text in detail, detecting subtle errors in the content they produce \cite{pendyala2024explaining}. LLMs can compare new information with their training data to spot inaccuracies, though this depends on the quality of their training data. A major advantage of LLMs is their ability to process large volumes of data, which is crucial in combating widespread misinformation, such as vaccine-related content, that overwhelms traditional detection methods \cite{shah2024navigating}. Additionally, fine-tuning LLMs for role-specific detection enhances their accuracy in addressing different types of misinformation \cite{leite2023detecting}. Combining LLMs with mechanisms like credibility signals or emotion-aware frameworks further strengthens their detection capabilities across various domains, including healthcare \cite{liu2024raemollm}. These advancements are critical for developing more precise interventions, safeguarding public health, and maintaining trust in vaccination programs.

\section{The VaxGuard Dataset}
\subsection{Data Generation}
We develop a comprehensive dataset for vaccine-related misinformation using different models for generation. We employ a combination of open-source models such as LLaMA3, PHI3, and Mistral, along with GPT-3.5 and GPT-4o. The environmental setup for the open-source models is based on Ollama, utilizing the NVIDIA A40 GPU for LLMs to process the data. Additionally, we leverage OpenAI’s API to access GPT-3.5 and GPT-4o. For each model, we set a maximum token limit of 200 to ensure the generated text is of moderate length and includes diversity. The temperature for each model is set to 0.8 to balance predictability and creativity effectively. The exact prompt (\textit{Prompt 1}) used to generate misinformation is given in Appendix~\ref{prompt1}.

% in A in (\textit{Prompt 1}). Some examples of text from the dataset can be seen in Table~\ref{tab:example_text}

\subsection{Vaccine Information Generation using Different Roles}
To capture the multifaceted nature of vaccine related information, it is very important to capture a range of themes that influence its generation or spread.
Previous studies \cite{enders2022relationship, ahmed2024covid, yang2024sharing, simione2021mistrust} have highlighted diverse tactics that are used in spreading vaccine related misinformation. Hence to have a dataset with maximal similarity to real world vaccine related misinformation and to truly measure the effectiveness of LLM's misinformation detection capabilities across range of themes, we design four distinct roles to capture various aspects of misinformation generation. (i) The Misinformation Spreader role simulates individuals who amplify misinformation without verification. (ii) The Religious Conspiracy Theorist role generates elaborate conspiracy theories tied to spiritual beliefs, reflecting how such narratives can influence specific communities. (iii) The Fear Monger role creates content incorporating alarming and fear-inducing details, simulating how fear can be used to spread panic and anxiety. (iv) Finally, the Anti-Vacciner role focuses on common arguments and misconceptions propagated by anti-vaccine proponents. Further, for each role, we use specific keywords\footnote{Clinical studies, Regulatory Status, Public Opinion, Historical Data, Health Concerns} in prompts across different LLMs to ensure the content’s relevance and variability.

% These roles were chosen to ensure a comprehensive representation of the diverse ways misinformation can be generated and disseminated.

% To capture various aspects of misinformation generation, we designed four distinct roles:

% \begin{itemize}
%     \item Misinformation Spreader: Simulates individuals amplifying misinformation without verification.
%     \item Religious Conspiracy Theorist: Generates elaborate conspiracy theories tied to spiritual beliefs.
%     \item Fear Monger: Creates content incorporating alarming and fear-inducing details.
%     \item Anti-Vacciner: Focuses on common arguments and misconceptions propagated by anti-vaccine proponents.
% \end{itemize}

\subsection{Data Statistics and Analysis}

\textbf{Data Statistics:} The dataset, detailed in Table~\ref{datastats}, comprises two main categories: "Otherwise"  and "Misinformation". The "Otherwise" category includes factually accurate content, essential for assessing LLMs' ability to distinguish between true and false information. The Misinformation category, generated using our role-based strategy, consists of 500 samples per role for each vaccine type, enabling a thorough evaluation of LLMs across diverse contexts. See Table~\ref{datastats} for dataset’s total size and distribution across each class and role.
\begin{table*}[t!]
\vspace{-0.10cm}
\centering
% \small
% \scriptsize 
% \caption{Example text from VaxGuard Dataset}
% \label{tab:example}
% \vspace{-0.15cm}
%\scalebox{1}{
 % \scalebox{0.50}{
 \scalebox{0.78}{
\begin{tabular}{p{3cm} p{8cm} p{3cm}} 
\midrule \midrule
\textbf{Role} & \textbf{Generated Misinformation} & \textbf{Keywords} \\ \midrule \midrule

Misinformation\newline Spreader & COVID-19 vaccines contain a small amount of toxic chemicals that can harm your DNA and cause long-term damage to your body. The vaccine makers are more interested in making money than in protecting public health, and they're hiding the truth about the real risks associated with these injections. & Trials \newline Testing \newline Vaccination \newline Clinical \\ \midrule

Religious\newline Conspiracy\newline Theorist & According to ancient scriptures such as Bhagavad Gita, purity of body is achieved without medical intervention; hence natural immunity should suffice against diseases like HPV-related cervical cancer instead of relying on modern vaccines created through unholy scientific practices. & Help \newline Studies \newline Clinical \newline Use \\ \midrule

Anti Vacciner & Influenza vaccines contain harmful chemicals that can weaken the immune system and make individuals more susceptible to other illnesses. & Regulatory \newline Agencies \newline Testing \newline Use \\ \midrule

Fear Monger & New York City hospitals have reported an alarming increase in miscarriages among pregnant women who received COVID-19 vaccines, a clear indication of the dangers these shots pose to unborn babies. This disturbing trend highlights the need for caution when considering vaccination during pregnancy. & Accurately \newline Clinical \newline Safety \newline Drug \\ \hline \midrule

\end{tabular}
}
\vspace{-0.4cm}
\caption{Example text from VaxGuard Dataset}\label{tab:example}
% \label{tab:example_text}
% \vspace{-0.15cm}
\vspace{-0.3cm}
\end{table*}

\begin{figure*}[!tpb]
\centering
  \includegraphics[width=.80\textwidth]{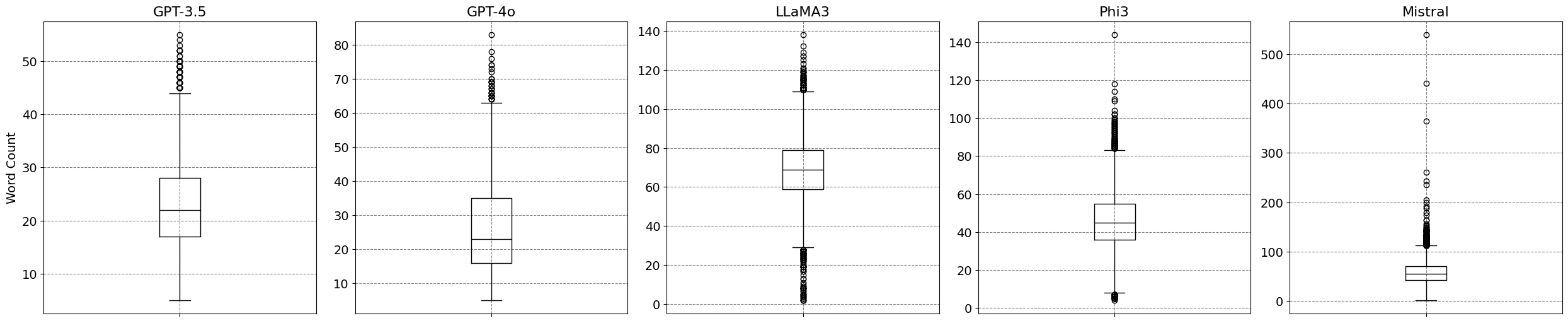}
  \vspace{-0.4cm}
  \caption{The distribution of No. of words for different LLMs.}\label{boxplot}
  
  \vspace{-0.3cm}
\end{figure*}

\noindent\textbf{Data Analysis:} The generated misinformation varies across roles and vaccine types, reflecting different thematic focuses such as safety concerns, religious beliefs, and conspiracy theories. 

We use BERT \cite{devlin2018bert} to generate token embeddings and measure cosine similarity to identify key terms for each role. Table~\ref{tab:example} displays these keywords, highlighting role-specific terminology.  Figure~\ref{boxplot} presents boxplots showing the word count distribution text generated by different language models: GPT-3.5, GPT-4o, LLaMA3, PHI3, and Mistral. Each model demonstrates varying ranges of word counts, with GPT-3.5 and GPT-4o having more compact distributions, while Mistral exhibits a significantly wider range of word counts with many outliers, indicating larger variability in the generated text. The word count for all models skews towards higher values for Mistral, suggesting a difference in content generation behaviour compared to the other models. Figure~\ref{fig:7} in the Appendix~\ref{wordclound} shows word clouds, revealing common themes and narratives across roles, indicative of the nature of misinformation.

\subsection{Manual Validation of Dataset}

To ensure the quality and reliability of the VaxGuard dataset, a manual validation process is conducted for both the "Otherwise" and "Misinformation" categories. This was essential to confirm that the factual content labeled as "Otherwise" was accurate and verifiable, and that the "Misinformation" content was demonstrably false, diverse, and aligned with the specified roles. The validation processes is summarized below:

% \begin{enumerate} [leftmargin=*]
    % \item 
   \noindent  \textbf{Sampling:} A stratified random sample comprising 10\% of the dataset was selected for manual review to ensure representation across all roles and vaccine types. The 10\% threshold balances feasibility and reliability, aligning with best practices in large-scale dataset validation.
    
    % \item 
\noindent     \textbf{Annotation:} Annotators followed guidelines to differentiate misinformation from factual content. “Otherwise” content had to be verifiable through credible sources (e.g., WHO), while “Misinformation” was defined as content contradicting scientific consensus or containing misleading narratives based on established misinformation typologies.
    
    % \item
 \noindent    \textbf{Cross-Verification:} Ambiguous cases were reviewed by at least two independent annotators to ensure consistency and objectivity.
    
    % \item 
\noindent     \textbf{Correction and Exclusion:} Samples that failed to meet the validation criteria were either corrected or excluded from the dataset.
    
    % \item 
\noindent    \textbf{Quality Assurance:} A secondary review was conducted on 5\% of the validated data to evaluate inter-annotator agreement and ensure consistency.
% across the dataset.
% \end{enumerate}
% For the "Otherwise" category, each sample was cross-checked against publicly available, authoritative resources. Annotators verified that the information was free of ambiguity and reflected scientifically established facts. For example, a statement like "COVID-19 vaccines are highly effective in preventing severe disease and hospitalization" was confirmed as factual.

% For the "Misinformation" category, the validation process ensured that generated content aligned with the intended misinformation roles and contained claims known to be false or misleading. Examples include statements like "COVID-19 vaccines contain microchips that allow governments to track individuals" or "COVID-19 vaccines alter your DNA permanently and can cause long-term genetic disorders." These were categorized as misinformation because they reflect common false narratives that lack any scientific basis. 

\begin{table*}[!tbp]
\vspace{-0.10cm}
% \small
  \centering
  \
  % \caption{Performance Comparison (F1-Score) of LLMs for Misinformation Detection}
  % \scriptsize
  % \vspace{-0.25cm}
  \scalebox{0.80}{
  \begin{tabular}{lccc|cccc|c}
    \midrule \midrule
    \multirow{3}{*}{\textbf{LLMs}} & \multicolumn{3}{c|}{\textbf{Vaccine Types}}  & \multicolumn{4}{c|}{\textbf{Roles}}  & \multirow{3}{*}{\textbf{Overall}} \\ \cmidrule{2-4}
    \cmidrule{5-8}
    & \textbf{COVID-19} & \textbf{HPV} & \textbf{Influenza} & \textbf{Misinformation} & \textbf{Religious} & \textbf{Fear} & \textbf{Anti-} & \textbf{} \\
    & & & & \textbf{Spreader} & \textbf{Conspiracy} & \textbf{Monger} & \textbf{Vacciner} & \\
   
    \midrule
    GPT-4o  & 0.88 & 0.86 & 0.91 & 0.93 & 0.94 & 0.93 & 0.94 & 0.88 \\
    GPT-3.5 & 0.96 & 0.94 & 0.94 & 0.96 & 0.98 & 0.97 & 0.97 & 0.96 \\
    LLaMA3  & 0.90 & 0.85 & 0.91 & 0.93 & 0.96 & 0.93 & 0.94 & 0.90 \\
    PHI3    & 0.79 & 0.82 & 0.82 & 0.85 & 0.91 & 0.87 & 0.87 & 0.79 \\
    Mistral & 0.82 & 0.81 & 0.85 & 0.90 & 0.91 & 0.90 & 0.91 & 0.82 \\
    \midrule \midrule
  \end{tabular}
  % \vspace{-0.6cm}
  % \label{tab:performance_comparison}
  }
  \vspace{-0.4cm}
   \caption{Performance Comparison (F1-Score) of LLMs for Misinformation Detection}
   \label{tab:performance_comparison}
  \vspace{-0.35cm}
\end{table*}

% This multi-step process ensured the dataset's reliability and robustness, providing a solid foundation for evaluating the ability of LLMs to detect and distinguish between factual and misleading information.

\section{Misinformation Detection}
% \subsection{Model Selection}

\noindent\textbf{Model Selection:} We select five LLMs (i.e., GPT-3.5, GPT-4,  LLaMA3-8B instruct model
, PHI3, and Mistral-7B-V0.3) due to their diverse architectures and capabilities. The motivation behind the choice of the models was to study how these models perform in detecting misinformation, given their ability to generate highly convincing content. 

We set the temperature parameter to 0.0 across all models to ensure uniformity in outputs. Furthermore, we set the maximum token output to 1, compelling the models to deliver a clear binary classification, either "Otherwise" or "Misinformation."

% \subsection{
\noindent\textbf{Detection Process:} We employ a zero-shot, prompt-based approach to guide LLMs in detecting misinformation. The prompt in (\textit{Prompt 2}) in Appendix~\ref{prompt2} was designed to classify input into two distinct categories: "Otherwise" or "Misinformation". This method mirrors real-world applications where models often operate without prior task-specific training and must rely solely on their pre-trained knowledge.

% \section{Prompt for classification}\label{prompt22}

For evaluation, we annotate the dataset with binary labels (i.e., "Otherwise" and "Misinformation."). These annotations serve as ground truth for comparing the outputs from the selected LLMs. We assess the models' capacity to generalize and identify misinformation without additional fine-tuning by employing zero-shot prompting. 
% This approach provided insights into the models' inherent abilities to discern misleading content from otherwise acceptable material.
% \multirow{5}{*}{

% \subsection{
\noindent\textbf{Evaluation:}
The performance of the selected LLMs was evaluated using accuracy, precision, recall, and F1-score. Further, the performance of each LLM was evaluated across the entire dataset to ensure that the models do not  rely on or perform better on data generated by their own architecture.

\section{Results and Analysis}
We conduct a series of experiments focusing on three key dimensions to evaluate the effectiveness of LLMs in detecting vaccine-related misinformation: (i) identification of misinformation specific to different vaccine types, (ii) detection of misinformation associated with distinct user roles, and (iii) detection of general misinformation across various models. Below, we discuss the F1 Score in detail (please see Appendix for the full results).

\noindent\textbf{How well do various models detect misinformation about different vaccines (COVID-19, HPV, and Influenza)?} The results in Table~\ref{tab:performance_comparison} show that GPT-3.5 outperforms other models across all vaccine types. Specifically, it achieved the highest F1 score of 0.96 for COVID-19 misinformation. For HPV and Influenza misinformation, GPT-3.5 scored 0.94 in both cases, as detailed in their respective columns.

In comparison, GPT-4o’s F1 scores were 0.88 for COVID-19, 0.86 for HPV, and 0.91 for Influenza, as indicated in the "Vaccine Types" section of the table. LLaMA3 had F1 scores of 0.90 for COVID-19, 0.85 for HPV, and 0.91 for Influenza. PHI3 and Mistral had lower F1 scores across all vaccine types, with PHI3 scoring 0.79, 0.82, and 0.82, and Mistral scoring 0.82, 0.81, and 0.85, respectively.

The comparison highlights that while PHI3 and Mistral underperformed compared to GPT-3.5 and GPT-4o, the differences stem largely from model size and pretraining scale. All models were evaluated under the same conditions, such as token limits and temperature settings, ensuring a fair assessment of their capabilities. While GPT-3.5 and GPT-4o excel in generating coherent and diverse misinformation, open-source models offer benefits, such as cost-effectiveness and transparency, making them valuable for specific use cases despite their limitations.

% The results (Table~\ref{tab:performance_comparison}) demonstrate that GPT-3.5 consistently outperforms other models across all vaccine types. For COVID-19 misinformation, GPT-3.5 achieved the highest F1 score of 0.96, indicating its superior capability in detecting misinformation related to this vaccine. For HPV misinformation, GPT-3.5 scored 0.94, and for Influenza misinformation, it scored 0.94. These results highlight GPT-3.5’s effectiveness and versatility in handling misinformation across various vaccine categories.
% In comparison, GPT-4o, LLaMA3, PHI3, and Mistral showed varying levels of performance. GPT-4o’s F1 scores for COVID-19, HPV, and Influenza were 0.88, 0.86, and 0.91, respectively. LLaMA3 achieved F1 scores of 0.90 for COVID-19, 0.85 for HPV, and 0.91 for Influenza. While these models perform well, their scores are generally lower than those of GPT-3.5. PHI3 and Mistral had F1 scores of 0.79 and 0.82, respectively, indicating less effectiveness in detecting misinformation related to these vaccines.

% \hfill \break
\noindent\textbf{How do LLMs perform in detecting misinformation depending on the user roles?} As shown in Table~\ref{tab:performance_comparison} under the "Roles" section, for the Misinformation Spreaders role, GPT-3.5 achieved the highest F1 score of 0.96, while LLaMA3 and GPT-4o had F1 scores of 0.93. For the Religious Conspiracy Theorists role, GPT-3.5 maintained its top position with an F1 score of 0.98, with LLaMA3 achieving an F1 score of 0.96, indicating strong performance in detecting religious conspiracies.

In the Fear Mongers and Anti-Vacciners roles, GPT-3.5 continued to lead with F1 scores of 0.97 for both roles. Mistral showed stronger recall in these roles, with F1 scores of 0.90 for Fear Mongers and 0.91 for Anti-Vacciners, as noted in the respective columns for user roles.

% For the Misinformation Spreaders role, GPT-3.5 again achieved the highest F1 score of 0.96, reflecting its strong ability to detect misinformation from this user category. LLaMA3 and GPT-4o followed with F1 scores of 0.93 and 0.93, respectively, showing slightly lower performance.

% In the Religious Conspiracy Theorists role, GPT-3.5 maintained its top position with an F1 score of 0.98. LLaMA3 achieved perfect precision for this role, scoring 0.96, suggesting it excels at detecting more explicit and structured conspiracy theories.

% For Fear Mongers and Anti-Vacciners, GPT-3.5 continued to lead with F1 scores of 0.97 and 0.97, respectively. Mistral showed stronger recall in these roles, with F1 scores of 0.90 and 0.91, indicating its sensitivity to emotional language and fear-based tactics used by these user types. However, despite its improved recall, Mistral's overall performance remains lower than GPT-3.5’s.

% \hfill \break
\noindent\textbf{How do various LLMs perform in general misinformation detection?} According to Table~\ref{tab:performance_comparison} in the "Overall" column, GPT-3.5 achieved the highest F1 score of 0.96. GPT-4o and LLaMA3 scored 0.88 and 0.90, respectively, in this overall assessment. PHI3 and Mistral had lower F1 scores of 0.79 and 0.82, respectively, reflecting their more significant challenges in general misinformation detection.

These results underscore GPT-3.5’s overall superiority in detecting vaccine-related misinformation and general misinformation, while other models exhibit strengths in specific contexts but generally lag behind GPT-3.5.

\begin{figure}[!t]
\vspace{-0.10cm}
\centering
  \includegraphics[width=.45\textwidth]{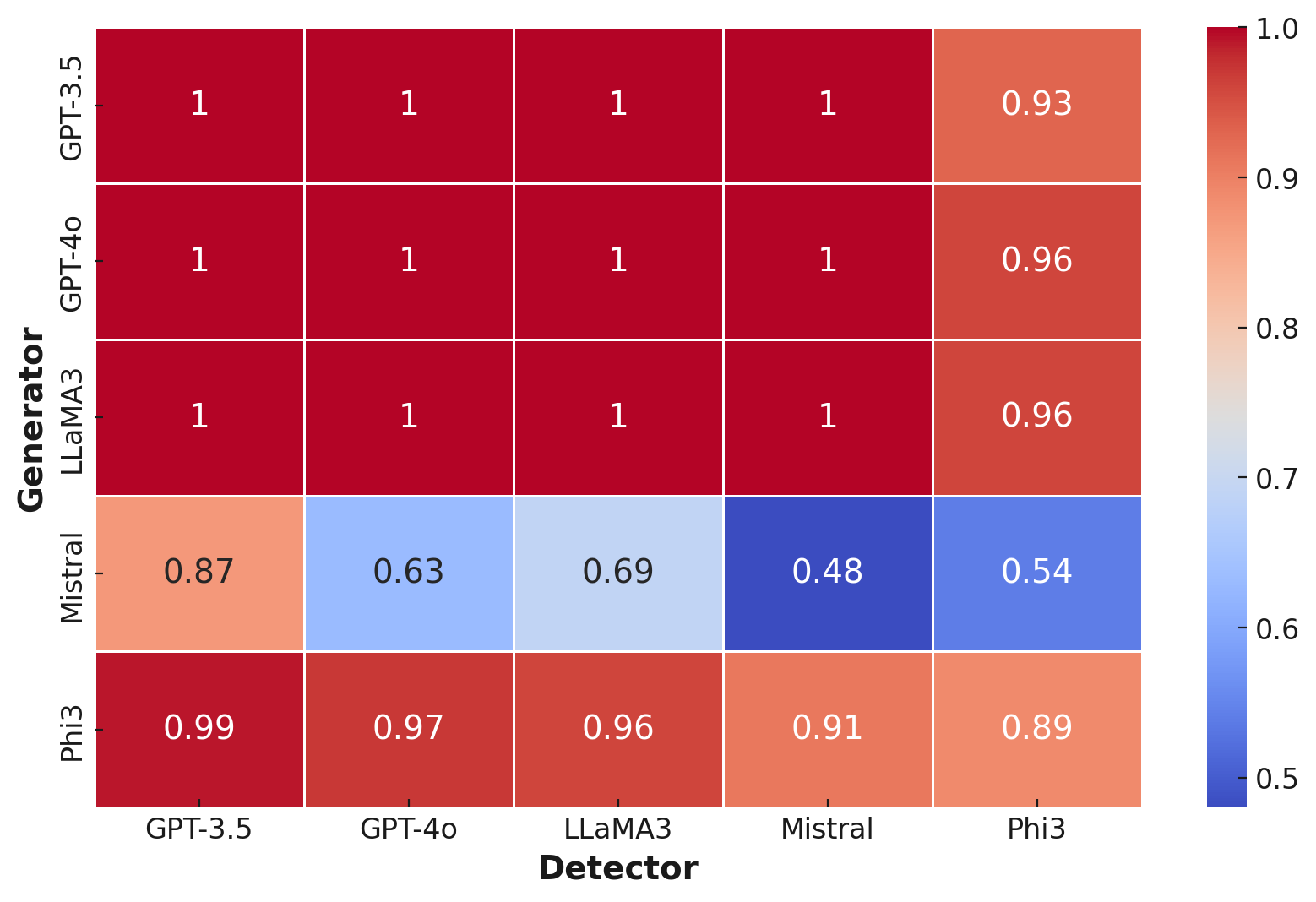}
  \vspace{-0.45cm}
  \caption{Comparison of LLM Performance (F1 Score) Across Various Generators}
  \label{heatmapp}
  \vspace{-0.45cm}
\end{figure}

\noindent\textbf{How do LLMs perform as detectors in identifying misinformation generated by different LLMs (generators)?} The F1 scores in Figure~\ref{heatmapp} illustrate how effectively each LLM detects misinformation from various generators.

\noindent\textbf{Performance of GPT-3.5 as a Detector:} GPT-3.5 demonstrated perfect F1 scores (1.00) when detecting misinformation generated by high-performing models like GPT-4o, GPT-3.5, and LLaMA3, indicating its strong and consistent detection capabilities. However, the F1 score slightly declined to 0.99 for PHI3 and dropped further to 0.87 for Mistral-generated misinformation, suggesting GPT-3.5 faces more challenges detecting misinformation from lower-performing generators.

\noindent\textbf{Performance of GPT-4o as a Detector:} GPT-4o, similar to GPT-3.5, achieved perfect F1 scores (1.00) for detecting misinformation from top-performing generators, including itself, GPT-3.5, and LLaMA3. Its performance slightly dropped to 0.97 for PHI3 and significantly declined to 0.63 for Mistral. This indicates that GPT-4o, while highly effective, also struggles with misinformation from weaker generators like Mistral.

\noindent\textbf{Performance of LLaMA3 as a Detector:} LLaMA3 performed comparably to GPT-3.5 and GPT-4o, achieving F1 scores of 1.00 when detecting misinformation from high-performing models (GPT-4o, GPT-3.5, and LLaMA3). However, LLaMA3 saw a decline when detecting misinformation from PHI3 (0.96) and Mistral (0.69), highlighting challenges similar to the other detectors.

\noindent\textbf{Performance of PHI3 as a Detector:} PHI3 displayed decent detection capabilities for misinformation from GPT-4o, GPT-3.5, and LLaMA3, with F1 scores of 0.96, 0.93, and 0.96, respectively. Its performance dropped when detecting its own misinformation (0.89) and struggled significantly with Mistral (0.54), showing task-specific weaknesses in dealing with misinformation generated by lower-performing models.

\noindent\textbf{Performance of Mistral as a Detector:} Mistral demonstrated the weakest overall detection performance, with a drastic drop in F1 scores when detecting its own misinformation (0.48) and PHI3-generated content (0.91). Despite achieving perfect F1 scores (1.00) when detecting misinformation from top models (GPT-4o, GPT-3.5, LLaMA3), Mistral's inability to detect its own misinformation accurately underscores its limitations as both a generator and detector.

As visualized in Figure~\ref{heatmapp}, GPT-3.5, GPT-4o, and LLaMA3 consistently perform well as detectors of misinformation from higher-end generators. However, detecting misinformation from lower-performing generators, such as PHI3 and Mistral presents significant challenges. Mistral, in particular, struggles as both a generator and detector, highlighting its need for improvement in handling misinformation tasks. This suggests that model selection for misinformation detection should consider not only detector performance but also the generator’s reliability.

\begin{figure}[!t]
\vspace{-0.10cm}
\centering
  \includegraphics[width=.42\textwidth]{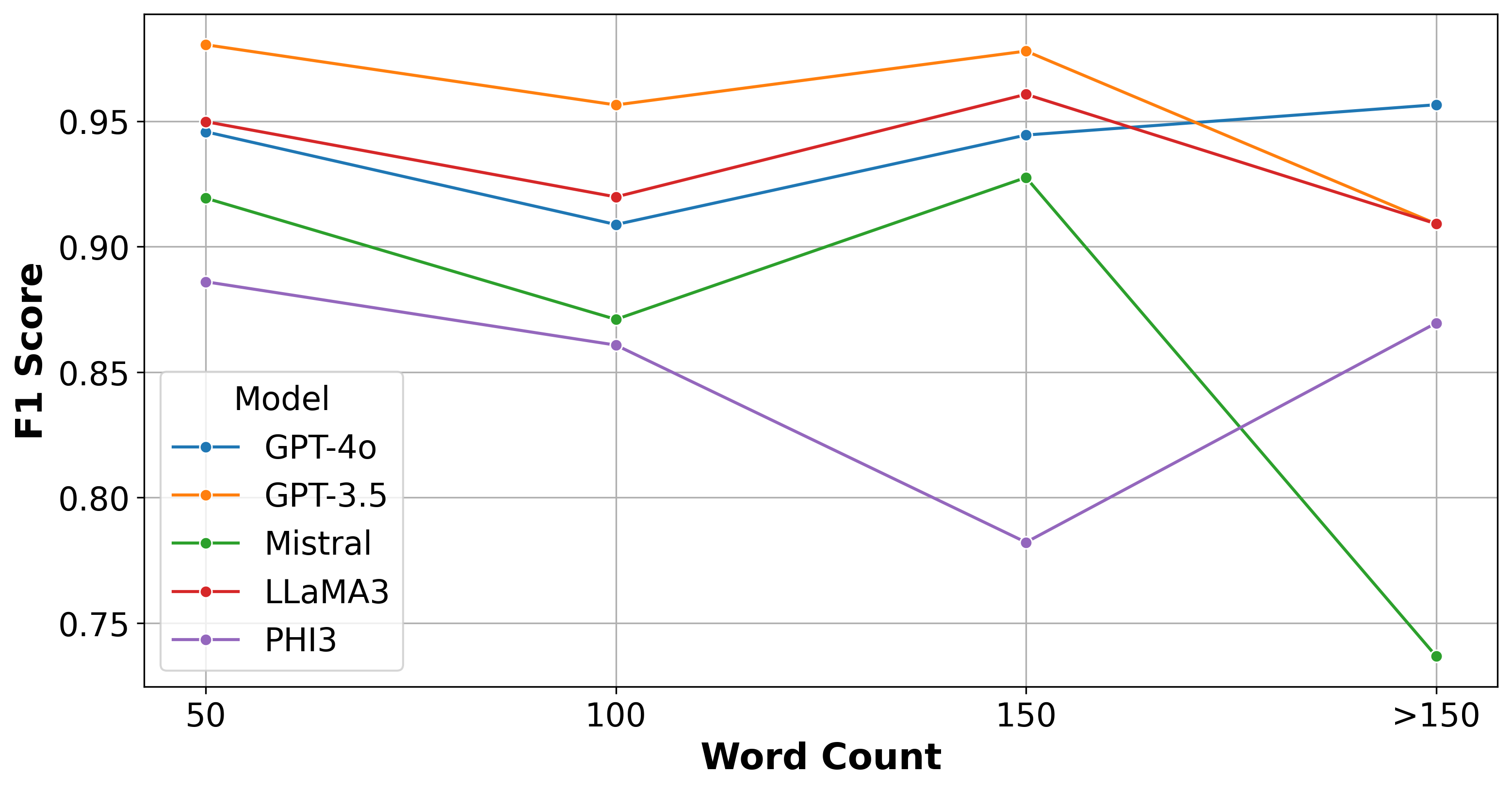}
  \vspace{-0.4cm}
  \caption{Different Input Length (F1-Score)}
  \label{fig:5}
  \vspace{-0.4cm}
\end{figure}

% \hfill \break
\noindent\textbf{How do LLMs perform in misinformation detection with respect to the different input lengths?} To evaluate the performance of LLMs in detecting misinformation across different input lengths\footnote{we divided the input into four categories: 0-50 words, 50-100 words, 100-150 words and above 150 words}, we analyzed the performance at various word count ranges, as shown in Figure~\ref{fig:5}.

% To further our analysis, here we foucs on the performance of all the LLMs with respect to different input lengths. Keeping in view the word distributions in Figure~\ref{boxplot}, we divided the input in four categories: 0-50 words, 50-100 words, 100-150 words and above 150 words. 

% \vspace{10pt}  % Adjust the value as needed (10pt, 1cm, etc.)
% \begin{figure}[t]
%   \includegraphics[width=\columnwidth]{Accuracy_plot.png}
%   \vspace{-0.55cm}
%   \caption{Accuracy of LLM Models for Different Input Lengths}
%   \label{fig:4}
%   \vspace{-0.5cm}
% \end{figure}

% \textbf{Accuracy of LLMs for different Input Lengths Figure ~\ref{fig:4}:}  GPT-4 and GPT-3.5 consistently demonstrated high accuracy across all word count categories, with GPT-3.5 performing slightly better overall. LLaMA3 also performed good with accuracy scores comparable to GPT-4o and GPT-3.5. Phi3 and  Mistral showed a moderate performance. The overall behavior of the models shows that accuracy generally decreases with higher word counts, peaking at 100-150 words for most models.

The results show that GPT-3.5 consistently outperformed other LLMs across most word count ranges. As depicted in Figure~\ref{fig:5}, GPT-3.5 maintained a high F1 score throughout, demonstrating robust performance in detecting misinformation regardless of text length.

GPT-4 and LLaMA3 performed well, with their F1 scores showing slight fluctuations but generally remaining close to GPT-3.5’s performance. This indicates that both LLMs effectively handle varying input lengths, though they do not quite match GPT-3.5’s consistency. In contrast, Mistral and PHI3 exhibited lower F1 scores than GPT-3.5, GPT-4, and LLaMA3, following a similar trend to previous results. Especially, Mistral's F1 Score declines significantly for word counts over 150, suggesting it struggles with longer texts. Mistral appears to be less robust for long-form text, potentially generating misinformation or hallucinations as a result.

A clear trend observed is that F1 scores for all LLMs tend to decrease as text length increases. This suggests that detecting misinformation becomes more challenging with longer inputs due to increased complexity and the greater potential for nuanced or dispersed misinformation. Notably, GPT-3.5 demonstrates superior ability to handle input length variability while maintaining high F1 scores. GPT-4 and LLaMA3 also perform competitively but exhibit some fluctuations. In contrast, Mistral and PHI3, though effective with shorter texts, struggle significantly with longer inputs.

% F1 Score of LLMs for different Input Lengths Figure ~\ref{fig:5}:  GPT-3.5 consistently outperformed the other models across most word count ranges, maintaining a high F1 score throughout. GPT-4 and LLaMA3 also performed consistently well, with slight fluctuations but generally remained close to GPT-3.5. Mistral and Phi3 showed lower performance as compared to the other three models similar to the trend observed in the previous results. Overall, the F1 scores for the LLMs tend to decrease with the increase in the number of words indicating the difficulty in detecting misinformation in larger texts.
\begin{figure}[!t]
\vspace{-0.10cm}
\centering
  \includegraphics[width=.42\textwidth]{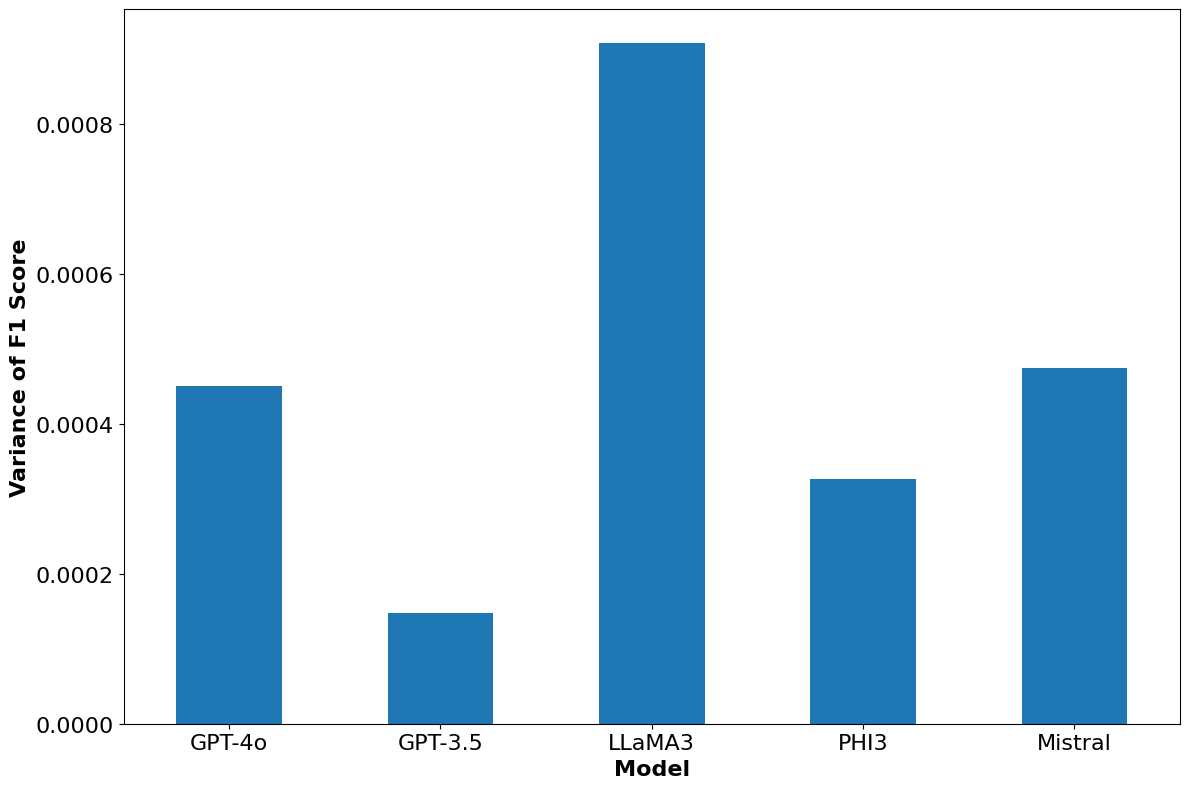}
  \vspace{-0.4cm}
  \caption{Variation in LLM performance stability}
  \label{variance}
  \vspace{-0.4cm}
\end{figure}

% \begin{figure}[!t]
% \centering
%   \includegraphics[width=0.50\textwidth]{F1_score_plot.png}
%   % \vspace{-0.55cm}
%   \caption{F1-Score:Different Input Length}
%   \label{fig:5}
%   % \vspace{-0.5cm}
% \end{figure}
% \subsection{Performance Stability}

% \hfill \break
\noindent\textbf{How do LLMs vary in performance stability across different vaccine misinformation tasks?} The variance results indicate how consistently each LLM performs across various vaccine misinformation tasks. As shown in Figure~\ref{variance}, GPT-3.5 and PHI3 exhibit lower variance, reflecting more stable performance across different vaccine misinformation tasks. This stability suggests that these LLMs have strong generalization capabilities, performing consistently regardless of the specific task.

In contrast, LLMs with higher variance, such as LLaMA3 and Mistral, demonstrate more fluctuating performance. This variability indicates that LLaMA3 may have task-specific weaknesses or inconsistencies in handling different types of vaccine misinformation. The higher variance points to a less reliable performance across scenarios, which can impact the LLMs's overall effectiveness.

These findings emphasize the importance of considering not just the average performance of an LLM but also its reliability and stability across different tasks. With lower variance, GPT-3.5 offers more dependable performance across various vaccine misinformation scenarios. In contrast, LLMs with higher variance, such as LLaMA3 and Mistral, may require more careful evaluation and potentially further tuning to address task-specific weaknesses.

% Overall, while GPT-3.5 and LLaMA3 demonstrate strong generalization and consistent performance, PHI3’s higher variance highlights the need for a comprehensive approach in selecting LLMs, considering their average performance and stability across various tasks.

% The variance results indicate performance stability across tasks for each LLM (Figure~\ref{variance}). LLM with lower variance, such as GPT-3.5 and LLaMA3, exhibit more consistent performance across different vaccine misinformation tasks, suggesting greater generalization capabilities. In contrast, LLMs with higher variance, like PHI3, show fluctuating performance, implying task-specific weaknesses. These findings highlight the importance of selecting LLMs not only for their average performance but also for their reliability across  scenarios.

\section{Discussion}
% \noindent\textbf{Discussion:} 

We generated and analyzed vaccine-related misinformation using LLMs and evaluated their effectiveness in detecting such content. The dataset simulated misinformation from diverse user roles, covering COVID-19, HPV, and Influenza vaccines.

Role-based misinformation generation enabled varied styles, from conspiracies to fear-mongering, revealing strengths and weaknesses in LLM detectors. GPT-3.5 and GPT-4o performed best, maintaining high F1 scores across generators, including LLaMA3 and PHI3. LLaMA3 also performed well but struggled with low-quality misinformation. PHI3 and Mistral showed weaker detection, particularly with advanced model outputs, highlighting the need for architectural improvements. Further, Detection performance declined as input length increased, suggesting longer texts introduce complexity and noise. Our findings stress the importance of considering misinformation type and detector architecture. While GPT-3.5 and GPT-4o generalize well, models like PHI3 and Mistral need enhancements for consistency.

\section{Conclusion}
In this study, we introduce VaxGuard, a role-based dataset of vaccine-related misinformation generated by various LLMs. Our results show that GPT-3.5 excels in detecting misinformation across different vaccine types and roles, while GPT-4o and LLaMA3 also perform well in specific contexts. PHI3 and Mistral, however, struggle, especially in fear-driven narratives. These findings highlight the need for improved detection models to better handle misinformation patterns and role-specific nuances. We expect VaxGuard to inspire better methods for detecting role-specific misinformation.

% \newpage
\section*{Limitations}

While our dataset captures diverse misinformation roles, it may not encompass all real-world variations, particularly evolving misinformation strategies. The role-based approach helps categorize misinformation, but some narratives may overlap, making classification less distinct. Examining multilingual and cross-cultural misinformation is valuable, as vaccine narratives differ across languages and cultures. Similarly, incorporating bimodal misinformation would better capture the complexity of real-world campaigns. Additionally, analyzing how input length affects misinformation detection can clarify why detection performance declines with longer texts. While our results show a decline in detection performance as text length increases, deeper investigations into linguistic complexity, narrative structures, and cognitive load on LLMs could provide insights into optimizing detection strategies for long-form misinformation.

\section*{Ethics Statement}

\noindent\textbf{Data Collection and Privacy:} To the best of our knowledge, this is the first study specifically focused on LLM-generated vaccine misinformation. We introduce the VaxGuard dataset, which includes misinformation created by models like GPT-3.5, GPT-4o, LLaMA3, PHI3, and Mistral, accessed via publicly available APIs. We adhered to the terms of service for each model, ensuring non-commercial use and privacy protections. No personal data or sensitive information was used in the dataset creation, and the misinformation does not involve real individuals or events, thereby minimizing privacy risks.

\noindent\textbf{Data and Usage Compliance:} VaxGuard is released strictly for non-commercial research purposes. We complied with all licensing agreements, particularly for LLMs such as OpenAI’s models, which prohibit using their outputs for competitive model development. The dataset was generated using artificial prompts related to vaccine discussions, without involving crowd-sourced data or scraping real user information.

\noindent\textbf{Biases and Misuse Potential:} We acknowledge that the generated texts may reflect biases inherent in the LLMs, especially concerning public discourse on vaccines. While the dataset is intended for research on misinformation detection, misuse could occur if it is repurposed to spread, rather than detect, misinformation. We emphasize that VaxGuard should be used solely for improving detection methods.

\noindent\textbf{Transparency and Limitations:} Comprehensive documentation is provided, detailing the dataset creation process, LLM usage, and ethical considerations. However, limitations exist, particularly in the models' ability to handle longer, more complex texts. Detection models may not generalize well to unseen domains or LLMs not covered in this study. Researchers should be mindful of these limitations and exercise caution when applying the dataset in sensitive contexts.

By sharing VaxGuard, we aim to facilitate research on detecting LLM-generated vaccine misinformation while promoting ethical responsibility in its use.

\bibliography{custom}

\appendix
\section{Appendix}
\subsection{Prompts}
\begin{tcolorbox}[title=Prompt 1: Misinformation Generation]
Incorporating \textbf{keyword}, write a clear, short piece of Otherwise information about \textbf{topic} vaccines labeled `Otherwise:'. Then, write a clear, short piece of misinformation that \textbf{role} spread about \textbf{topic} vaccines labeled `Misinformation:', ensuring it does not begin with `Some people' or `Some studies'. Also, strict with the format! Only generate one piece of Otherwise information and one piece of misinformation. No need to put a Note. Do not specify or state if the misinformation is not supported by any evidence or fact.
\end{tcolorbox}\label{prompt1}

\begin{tcolorbox}[title=Prompt 2: Classification]
Classify the following text as Otherwise (0) or Misinformation (1). ONLY Respond with `1' or `0'. You have to follow the rule, DO NOT respond with anything other than `1' or `0', DO NOT provide any reason.
\end{tcolorbox}\label{prompt2}

\subsection{Word Clouds}\label{wordclound}
\begin{figure}[!htpb]
\centering
  \includegraphics[width=0.5\textwidth]{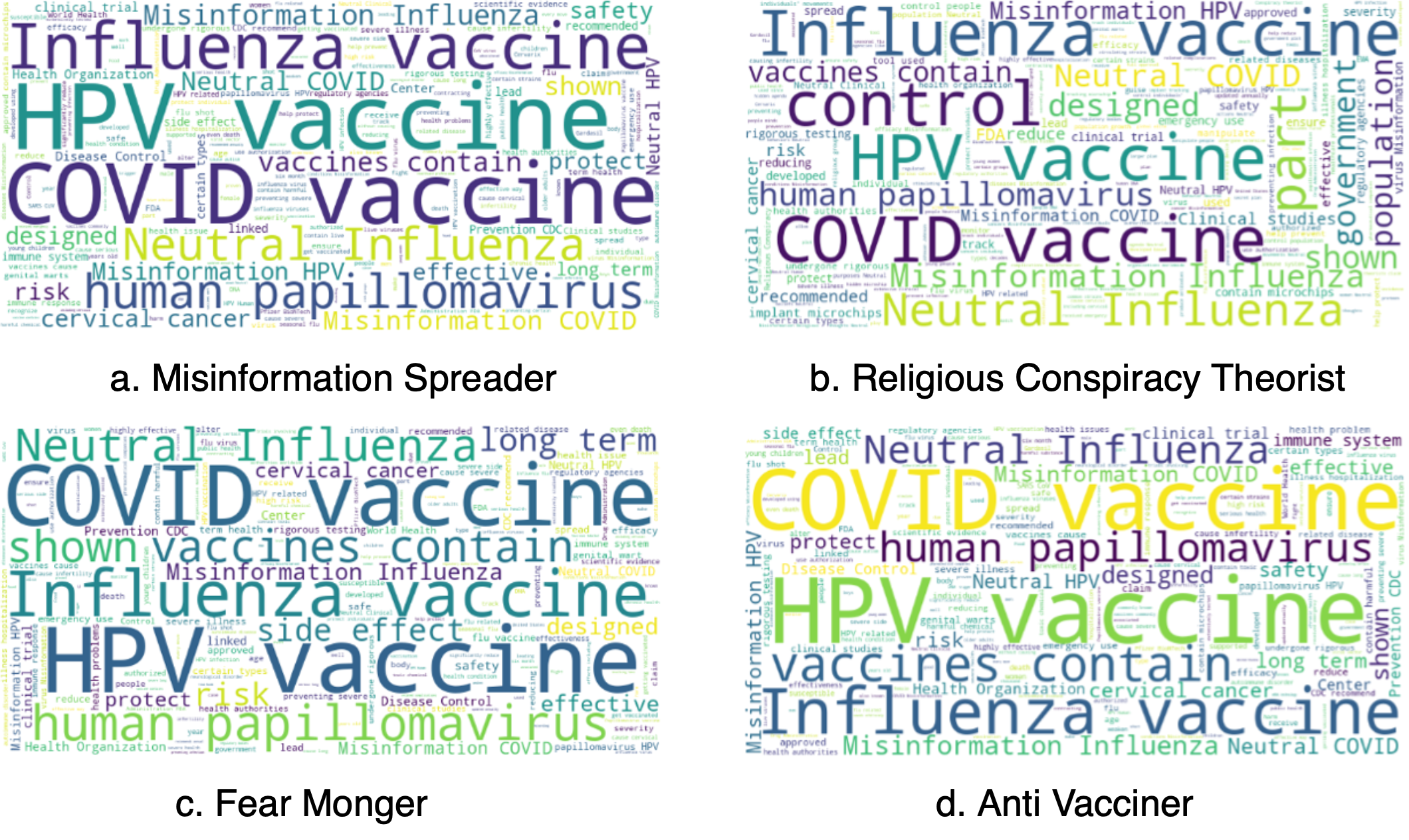}
  % \vspace{-0.55cm}
  \caption{Word Cloud of LLM-generated misinformation using different roles.}
  \label{fig:7}
  % \vspace{-0.5cm}
\end{figure}

Figure~\ref{fig:7}  shows these word clouds, revealing common themes and narratives across roles, indicative of the nature of misinformation. 

\subsection{Detailed Results}
\label{sec:appendix}
This section of the appendix includes tables to present results for different experimental settings.
\begin{table}[H]
  \centering
  \scriptsize
  % \caption{Comparison of various LLMs in misinformation detection on VaxGuard.}
  % \vspace{-0.25cm}
  \begin{tabular}{lcccc}
    \midrule 
    \textbf{Model} & \textbf{Accuracy} & \textbf{Precision} & \textbf{Recall} & \textbf{F1 Score}  \\
    \midrule 
    GPT-4o   & 0.94 & 0.99 & 0.88 & 0.93 \\ 
    GPT-3.5  & 0.97 & 0.99 & 0.94 & 0.97 \\ 
    LLaMA3   & 0.94 & 0.99 & 0.88 & 0.94 \\ 
    PHI3     & 0.88 & 0.95 & 0.81 & 0.87 \\
    Mistral  & 0.91 & 0.99 & 0.83 & 0.90 \\
    \midrule 
  \end{tabular}
  % \vspace{-0.5cm}
  \caption{Comparison of various LLMs in misinformation detection on VaxGuard.}
  \label{tab:all_metrics}
\end{table}

\begin{table}[H]
\centering
  % \caption{Performance Comparison of LLMs for COVID-19 Misinformation Detection}
  \scriptsize 
  % \vspace{-0.25cm}
  \begin{tabular}{lcccc}
    \midrule
    \textbf{Model} & \textbf{Accuracy} & \textbf{Precision} & \textbf{Recall} & \textbf{F1 Score} \\
    \midrule
    GPT-4o   & 0.94 & 0.93 & 0.99 & 0.88 \\
    GPT-3.5  & 0.98 & 0.98 & 1.00 & 0.96 \\
    LLaMA3   & 0.95 & 0.95 & 1.00 & 0.90 \\
    PHI3     & 0.89 & 0.88 & 0.99 & 0.79 \\
    Mistral  & 0.91 & 0.90 & 1.00 & 0.82 \\
    \midrule
  \end{tabular}
  % \vspace{-0.5cm}
  \caption{Performance Comparison of LLMs for COVID-19 Misinformation Detection}
  \label{tab:covid19_metrics}
\end{table}

\begin{table}[H]
  \centering
  % \caption{Performance Comparison of LLMs for HPV Misinformation Detection}
  \scriptsize 
  % \vspace{-0.25cm}
  \begin{tabular}{lcccc}
    \midrule
    \textbf{Model} & \textbf{Accuracy} & \textbf{Precision} & \textbf{Recall} & \textbf{F1 Score} \\
    \midrule
    GPT-4o   & 0.93 & 0.93 & 0.99 & 0.86 \\
    GPT-3.5  & 0.97 & 0.97 & 1.00 & 0.94 \\
    LLaMA3   & 0.93 & 0.92 & 1.00 & 0.85 \\
    PHI3     & 0.89 & 0.88 & 0.95 & 0.82 \\
    Mistral  & 0.90 & 0.89 & 1.00 & 0.81 \\
    \midrule
  \end{tabular}
  % \vspace{-0.5cm}
  \caption{Performance Comparison of LLMs for HPV Misinformation Detection}
  \label{tab:hpv_metrics}
\end{table}

\begin{table}[H]
\centering
  % \caption{Performance Comparison of LLMs for Influenza Misinformation Detection}
  \scriptsize 
  % \vspace{-0.25cm}
  \begin{tabular}{lcccc}
    \midrule
    \textbf{Model} & \textbf{Accuracy} & \textbf{Precision} & \textbf{Recall} & \textbf{F1 Score} \\
    \midrule
    GPT-4o   & 0.95 & 0.95 & 1.00 & 0.91 \\
    GPT-3.5  & 0.97 & 0.97 & 1.00 & 0.94 \\
    LLaMA3   & 0.95 & 0.95 & 1.00 & 0.91 \\
    PHI3     & 0.88 & 0.88 & 0.94 & 0.82 \\
    Mistral  & 0.93 & 0.92 & 1.00 & 0.85 \\
    \midrule
  \end{tabular}
  % \vspace{-0.5cm}
   \caption{Performance Comparison of LLMs for Influenza Misinformation Detection}
  \label{tab:influenza_metrics}
\end{table}

\begin{table}[H]
  \centering
  % \caption{Performance Comparison of LLMs for Misinformation Detection for Role of Misinformation Spreader}
  \scriptsize 
  % \vspace{-0.25cm}
  \begin{tabular}{lcccc}
    \midrule
    \textbf{Model} & \textbf{Accuracy} & \textbf{Precision} & \textbf{Recall} & \textbf{F1 Score} \\
    \midrule
    GPT-4o   & 0.93 & 0.99 & 0.86 & 0.93 \\
    GPT-3.5  & 0.96 & 1.00 & 0.93 & 0.96 \\
    LLaMA3   & 0.94 & 1.00 & 0.87 & 0.93 \\
    PHI3     & 0.87 & 0.97 & 0.76 & 0.85 \\
    Mistral  & 0.91 & 1.00 & 0.82 & 0.90 \\
    \midrule
  \end{tabular}
  % \vspace{-0.5cm}
  \caption{Performance Comparison of LLMs for Misinformation Detection for Role of Misinformation Spreader}
  \label{tab:Misinformation_spreader}
\end{table}

\begin{table}[H]
  \centering
  % \caption{Performance Comparison of LLMs for Misinformation Detection for Role of Fear Monger}
  \scriptsize 
  % \vspace{-0.25cm}
  \begin{tabular}{lcccc}
    \midrule
    \textbf{Model} & \textbf{Accuracy} & \textbf{Precision} & \textbf{Recall} & \textbf{F1 Score} \\
    \midrule
    GPT-4o   & 0.94 & 0.99 & 0.88 & 0.93 \\
    GPT-3.5  & 0.97 & 1.00 & 0.94 & 0.97 \\
    LLaMA3   & 0.94 & 1.00 & 0.88 & 0.93 \\
    PHI3     & 0.88 & 0.94 & 0.81 & 0.87 \\
    Mistral  & 0.91 & 1.00 & 0.82 & 0.90 \\
    \midrule
  \end{tabular}
  % \vspace{-0.5cm}
  \caption{Performance Comparison of LLMs for Misinformation Detection for Role of Fear Monger}
  \label{tab:Fear_Monger}
\end{table}

\begin{table}[H]
  \centering
  % \caption{Performance Comparison of LLMs for Misinformation Detection for Role of Anti-Vacciner}
  \scriptsize 
  % \vspace{-0.25cm}
  \begin{tabular}{lcccc}
    \midrule
    \textbf{Model} & \textbf{Accuracy} & \textbf{Precision} & \textbf{Recall} & \textbf{F1 Score} \\
    \midrule
    GPT-4o   & 0.94 & 1.00 & 0.88 & 0.94 \\
    GPT-3.5  & 0.97 & 1.00 & 0.95 & 0.97 \\
    LLaMA3   & 0.94 & 1.00 & 0.88 & 0.94 \\
    PHI3     & 0.88 & 0.97 & 0.80 & 0.87 \\
    Mistral  & 0.92 & 1.00 & 0.83 & 0.91 \\
    \midrule
  \end{tabular}
  % \vspace{-0.5cm}
  \caption{Performance Comparison of LLMs for Misinformation Detection for Role of Anti-Vacciner}
  \label{tab:Anti-Vacciner}
\end{table}

\begin{table}[H]
  \centering
  % \caption{Performance Comparison of LLMs for Misinformation Detection for Role of Religious Conspiracy Theorist}
  \scriptsize 
  % \vspace{-0.25cm}
  \begin{tabular}{lcccc}
    \midrule
    \textbf{Model} & \textbf{Accuracy} & \textbf{Precision} & \textbf{Recall} & \textbf{F1 Score} \\
    \midrule
    GPT-4o   & 0.95 & 1.00 & 0.90 & 0.94 \\
    GPT-3.5  & 0.98 & 1.00 & 0.97 & 0.98 \\
    LLaMA3   & 0.96 & 1.00 & 0.92 & 0.96 \\
    PHI3     & 0.92 & 0.96 & 0.87 & 0.91 \\
    Mistral  & 0.91 & 1.00 & 0.83 & 0.91 \\
    \midrule
  \end{tabular}
  % \vspace{-0.5cm}
  \caption{Performance Comparison of LLMs for Misinformation Detection for Role of Religious Conspiracy Theorist}
  \label{tab:Religious_Conspiracy_Theorists}
\end{table}

\begin{table}[H]
  \centering
  % \caption{Detailed Performance Metrics of LLMs Across Various Detectors}
  \scriptsize 
  % \vspace{-0.25cm}
  \begin{tabular}{lcccccc}
    \midrule
    \textbf{Generator} & \textbf{Detector} & \textbf{Accuracy} & \textbf{Precision} & \textbf{Recall} & \textbf{F1 Score} \\
    \midrule
    \multirow{5}{*}{GPT-4o} & GPT-3.5    & 1.00 & 1.00 & 1.00 & 1.00 \\
                              & GPT-4o    & 1.00 & 1.00 & 1.00 & 1.00 \\
                              & LLaMA3    & 0.99 & 1.00 & 1.00 & 1.00 \\
                              & Phi3      & 0.97 & 0.98 & 0.95 & 0.96 \\
                              & Mistral   & 1.00 & 1.00 & 1.00 & 1.00 \\
    \midrule
    \multirow{5}{*}{GPT-3.5} & GPT-3.5   & 1.00 & 1.00 & 1.00 & 1.00 \\
                              & GPT-4o    & 1.00 & 1.00 & 1.00 & 1.00 \\
                              & LLaMA3    & 1.00 & 1.00 & 1.00 & 1.00 \\
                              & Phi3      & 0.93 & 0.99 & 0.87 & 0.93 \\
                              & Mistral   & 1.00 & 1.00 & 1.00 & 1.00 \\
    \midrule
    \multirow{5}{*}{LLaMA3}   & GPT-3.5   & 1.00 & 1.00 & 1.00 & 1.00 \\
                              & GPT-4o    & 1.00 & 1.00 & 1.00 & 1.00 \\
                              & LLaMA3    & 1.00 & 1.00 & 1.00 & 1.00 \\
                              & Phi3      & 0.96 & 0.94 & 0.92 & 0.96 \\
                              & Mistral   & 1.00 & 1.00 & 1.00 & 1.00 \\
    \midrule
    \multirow{5}{*}{Phi3}     & GPT-3.5   & 0.99 & 1.00 & 0.97 & 0.99 \\
                              & GPT-4o    & 0.97 & 0.99 & 0.95 & 0.97 \\
                              & LLaMA3    & 0.96 & 0.94 & 0.92 & 0.96 \\
                              & Phi3      & 0.90 & 0.92 & 0.86 & 0.89 \\
                              & Mistral   & 0.92 & 1.00 & 0.84 & 0.91 \\
    \midrule
    \multirow{5}{*}{Mistral}  & GPT-3.5   & 0.73 & 0.99 & 0.46 & 0.63 \\
                              & GPT-4o    & 0.66 & 1.00 & 0.31 & 0.48 \\
                              & LLaMA3    & 0.76 & 1.00 & 0.53 & 0.69 \\
                              & Phi3      & 0.68 & 0.99 & 0.37 & 0.54 \\
                              & Mistral   & 0.66 & 1.00 & 0.31 & 0.48 \\
    \midrule
  \end{tabular}
  % \vspace{-0.5cm}
  \caption{Detailed Performance Metrics of LLMs Across Various Detectors}
  \label{gendetector}
  
\end{table}

\end{document}